\documentclass[10pt,twocolumn,letterpaper]{article}

\usepackage{cvpr}              

\usepackage[dvipsnames]{xcolor}

\definecolor{cvprblue}{rgb}{0.21,0.49,0.74}
\usepackage[pagebackref,breaklinks,colorlinks,citecolor=cvprblue]{hyperref}

\def\paperID{*****} 
\def\confName{CVPR}
\def\confYear{2024}

\title{PCIE\_EgoHandPose Solution for EgoExo4D Hand Pose Challenge}

\author{
Feng Chen\\
Lenovo Research\\
{\tt\small chenfeng13@lenovo.com}
\and
Ling Ding \\
Lenovo Research\\
{\tt\small dingling3@lenovo.com}
\and
Kanokphan Lertniphonphan \\
Lenovo Research\\
{\tt\small klertniphonp@lenovo.com}
\and
Jian Li\\
Lenovo Research\\
{\tt\small lijian30@lenovo.com}
\and
Kaer Huang\\
Lenovo Research\\
{\tt\small huangke1@lenovo.com}
\and
Zhepeng Wang \\
Lenovo Research\\
{\tt\small wangzpb@lenovo.com}
}

\begin{document}
\maketitle
\begin{abstract}

This report presents our team's 'PCIE\_EgoHandPose' solution for the EgoExo4D Hand Pose Challenge at CVPR2024. The main goal of the challenge is to accurately estimate hand poses, which involve 21 3D joints, using an RGB egocentric video images provided for the task. This task is particularly challenging due to the subtle movements and occlusions. To handle the complexity of the task, we propose the Hand Pose Vision Transformer (HP-ViT). The HP-ViT comprises a ViT backbone and transformer head to estimate joint positions in 3D, utilizing MPJPE and RLE loss function. Our approach achieved the 1$^{st}$ position in the Hand Pose challenge with 25.51 MPJPE and 8.49 PA-MPJPE. Code is available at \url{https://github.com/KanokphanL/PCIE_EgoHandPose}

\end{abstract}

\section{Introduction}
\label{sec:introduction}

Recently, an egocentric video captured using a wearable camera has gained significant importance in the fields of human-computer interaction and robotics. The EgoExo4D Hand Pose challenge, part of the Ego4D Pose benchmark, brings attention to hand joints estimation from the egocentric perspective \cite{Grauman2023EgoExo4DUS}.

Various approaches \cite{Zheng2023POTTERPA, xu2022vitpose, li2021rle} have been proposed to estimate human bode pose. Pooling Attention Transformer (POTTER) \cite{Zheng2023POTTERPA}, based on transformer architectureserved as the baseline in this challenge. In \cite{xu2022vitpose} utilizes a vision transformer as backbone for feature extraction.

\begin{figure*}[t]
  \centering
   \includegraphics[width=0.9\linewidth]{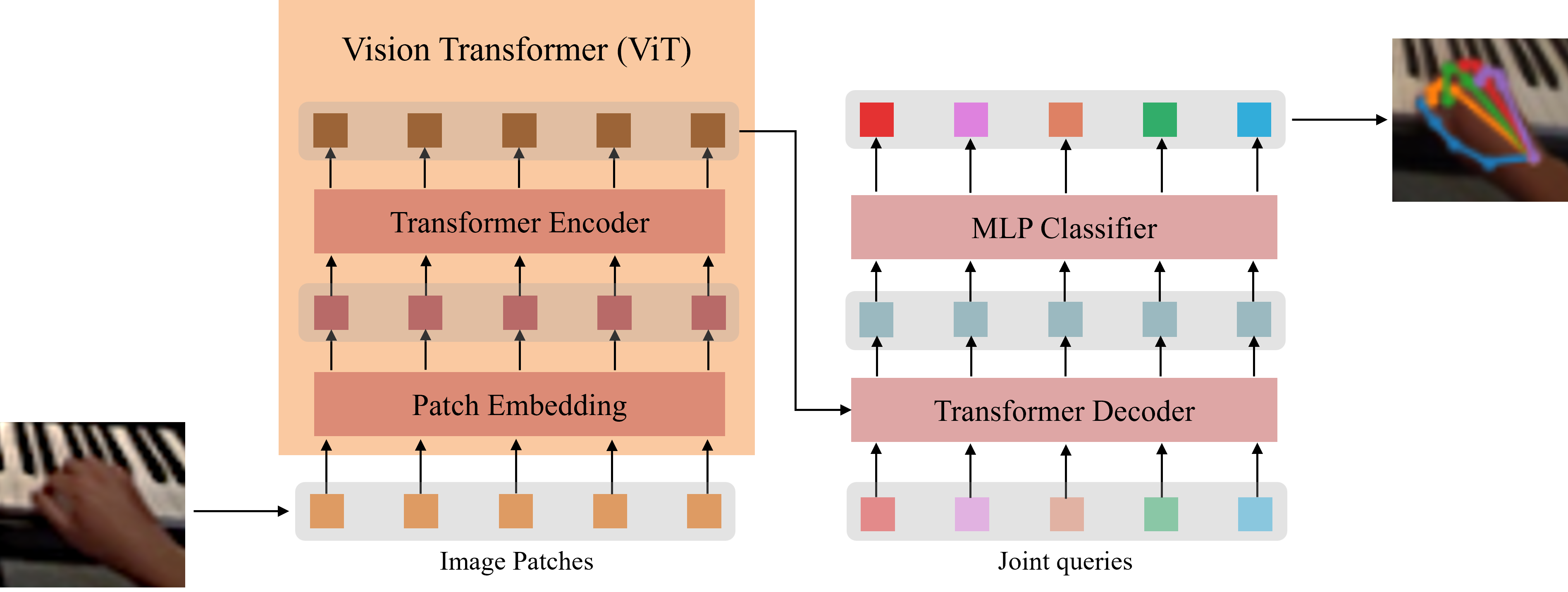}
   \caption{
    The framework of HP-ViT.
   }
   \label{fig:overview}
\end{figure*}

Our Hand Pose Vision Transformer (HP-ViT) incorporates the ViT backbone with a transformer head for estimating joint position in 3D, as shown in Figure \ref{fig:overview}. We mainly used MJPJE loss to train models. Additionally, we employed Residual Log-likelihood Estimation (RLE) loss \cite{li2021rle} to fine-tune the final model in each training scenario.

\section{Method}
\label{sec:method}

\subsection{Hand Pose Vision Transformer (HP-ViT)}

Visual Transformer (ViT) was utilized as the backbone \cite{xu2022vitpose} to extract features from the provided hand image. The input hand image $\mathcal{I} \in \mathcal{R}^{H\ast W\ast 3}$ is divided into patches $\mathcal{I}_p \in \mathcal{R}^{(H/16)\ast(W/16)\ast3}$ of 16 for input into a patch embedding layer. The output dimension is 21, corresponding to the number of hand joints. Leveraging the scalability properties of the ViT backbone, we experimented with ViT-Base, ViT-Large, and ViT-Huge \cite{Dosovitskiy2020Vit}. A standard transformer decoder \cite{Vaswani2017AttentionIA} was applied to locate keypoints. We used MLP classifier to output the coordinates of each joint.

\subsection{Regression Loss}

We employed regression loss based on RLE \cite{li2021rle} to minimize the gap between output and input distributions. In this work, we used Gaussian distribution as density function. In the training phase, we added a regression branch to the model to learn the parameters ${\mu}$ and ${\sigma}$ of the density function from the following equation.

\begin{equation}
    P(x|\mathcal{I}) = \frac{1}{{\sigma}\sqrt{2\pi}}e^{-\frac{(x-{\mu}^2)}{2{\sigma}^2}},
  \label{eq:mpjpe}
\end{equation}

$P(x|\mathcal{I})$ is the distribution of probability that the ground truth appears at $x$ given images $\mathcal{I}$.

After the parameterized process and residual log-likelihood estimation, the loss function consists of Gaussian distribution $G(\overline{x})=\mathcal{N}(0, I)$, the distribution $F(\overline{x})$ learned by the flow model, and constant $s$ defined as

\begin{equation}
    \mathcal{L}_{rle} = -logG(\overline{\mu}) - logF(\overline{\mu}) - log(s) + log({\sigma}),
  \label{eq:mpjpe}
\end{equation}

While $\mu_g$ is an observe label and $\overline{\mu} = (\mu_g-{\mu})/{\sigma}$.

\subsection{Test-Time Augmentation (TTA) and Ensemble}

During the validation and testing, we implemented TTA to enhance performance and decrease MPJPE. Specifically, only the vertical flip technique was utilized in TTA for this hand pose dataset.

Throughout the challenge, various methods, trained models, and outputs were explored in our experiments. Ensemble techniques were employed through \cite{roman2021weight} to merge the outputs. The fused joint coordinates were calculated as the weighted average of the merged joints. Optimal weighted parameters were determined through random search conducted on the validation dataset.


\section{Experiment}

\subsection{Dataset and Evaluation metric}

We conducted experiments using the Ego-Exo4D hand pose dataset \cite{Grauman2023EgoExo4DUS}. The dataset contains 68K and 340K manual annotations for 2D and 3D, respectively. Additionally, it includes 4.3M and 21M automatic annotations for 2D and 3D. Hand bounding boxes are also provided in the annotation. The Mean Per Joint Position Error (MPJPE) and Procrustes aligned Mean Per Joint Position Error (PA-MPJPE) are used for evaluation. Both metrics are evaluated in the context of the ego-centric coordinate system in millimeters (mm). 

\subsection{Implementation Detail and Results}

First, we evaluated a baseline method based on POTTER \cite{Zheng2023POTTERPA}.  Pre-trained models V1 and V2, trained on different annotated datasets, were tested as shown in \ref{tab: baseline}. Subsequently, we further trained the model with 70 epochs using the manual annotation dataset to verify the importance of the auto-annotated dataset. The PA-MPJPE of the manual and auto annotation pre-trained in the testing is quite close to the manual pre-trained model with a 0.01 difference. The later experimental training dataset will be based on a manual annotation dataset to save training time and resources. Moreover, we applied blur, median blur, and coarse dropout, random vertical flip, and YOLOX \cite{yolox2021} HSV random augmentations to the training data, enhancing  MPJPE and PA-MPJPE to 26.18 and 10.76, respectively, during validation.

We applied the ViT backbone to the baseline architecture led to improved the performance to 26.22 MPJPE and 10.01 PA-MPJPE. The ViT model was pre-trained using the MS COCO dataset \cite{Lin2014MicrosoftCC}. We trained the ViT model with the hand pose dataset for 70 epochs with a learning rate 5e-4, 64 batch size and MPJPE loss. A transformer head was then applied to further enhance performance. The best metric results from the validation set for various ViT backbones are presented in table \ref{tab:ViT size}. Results from ViT-huge were further improved with RLE loss and TTA as shown in table \ref{tab:ablation_test}. The model received further fine-tuning with RLE loss for an additional 20 epochs with a learning rate of 1e-4. Setting a low learning rate proved essential for stable training while employing RLE loss. TTA was also integrated during inference stage.

Finally, our HP-ViT results are summarized in Table \ref{tab: final}. The final results combined outputs from various ViT backbones and parameter configurations through weighted averages reduce MPJPE and PA-MPJPE by 0.20 and 0.14, respectively, compare to the single model HP-ViT in table \ref{tab:ablation_test}. Our approach achieved the 1$^{st}$ position in the Hand Pose challenge with a 25.51 MPJPE and 8.49 PA-MPJPE.

\begin{table*}
  \centering
  \begin{tabular*}{\textwidth}{@{\extracolsep{\fill}}
    lcccc} 
    \toprule
     &  \multicolumn{2}{c}{Validation}  &  \multicolumn{2}{c}{Test} \\
    \hline
     & MPJPE ($\downarrow$) & PA-MPJPE ($\downarrow$) & MPJPE ($\downarrow$) & PA-MPJPE ($\downarrow$)   \\
    \midrule
    V1 - Baseline with manual annotation  & -  & - & 30.57 & 11.14 \\
    V2 - Baseline with manual and auto annotation  & -  & - & 28.93 & 11.07 \\
    V1 fine-tuned with manual annotation & 28.68 & 10.97 & 30.07 & 10.75 \\
    V2 fine-tuned with manual annotation & 26.89 & 10.83 & 29.17 & 10.74 \\
    \bottomrule
  \end{tabular*}
  \caption{Performance of baseline \cite{Zheng2023POTTERPA} trained on the pre-trained model with different dataset}
  \label{tab: baseline}
\end{table*}

\begin{table*}
  \centering
  \begin{tabular*}{\textwidth}{@{\extracolsep{\fill}}
    lcccc} 
    \toprule
     &  \multicolumn{2}{c}{Validation}  &  \multicolumn{2}{c}{Test} \\
    \hline
     \multicolumn{1}{c}{Team} & MPJPE ($\downarrow$) & PA-MPJPE ($\downarrow$) & MPJPE ($\downarrow$) & PA-MPJPE ($\downarrow$)   \\
    \midrule
    PCIE\_EgoHandPose (Ours)  & \textbf{22.67}  & \textbf{8.59} & \textbf{25.51} & \textbf{8.49} \\
    Hand3D & - & - & 30.52 &9.30   \\
    Death Knight & - & - & 28.72 & 10.20 \\
    IRMV\_sjtu & - & - & 29.38 & 10.36\\
    Host\_1030\_Team (POTTER manual+auto v2) & - & - & 28.94 & 11.07 \\
    Host\_1030\_Team (POTTER manual v2) & - & - & 30.57 & 11.14 \\
    egoexo4d-hand (POTTER manual+auto v1) & - & - & 29.53 & 11.16 \\
    softbank-meisei (POTTER) & - & - & 62.14 & 19.85 \\
    \bottomrule
  \end{tabular*}
  \caption{EgoExo4D Hand Pose Challenge Leaderboard}
  \label{tab: final}
\end{table*}

\begin{table}
  \centering
  \begin{tabular}{lcc} 
    \toprule
     & MPJPE ($\downarrow$) & PA-MPJPE ($\downarrow$)   \\
    \midrule
    ViT-Base & 24.67 & 9.31\\
    ViT-Large & 23.38 & 9.06 \\
    ViT-Huge & 23.08 & 8.77 \\
    \bottomrule
  \end{tabular}
  \caption{Ablation Study on the different of ViT backbone in HP-ViT}
  \label{tab:ViT size}
\end{table}

\begin{table}
  \centering
  \begin{tabular}{lcc} 
    \toprule
     & MPJPE ($\downarrow$) & PA-MPJPE ($\downarrow$)   \\
    \midrule
    RLE Loss & 23.11 & 8.74 \\
    TTA & 22.87 & 8.73 \\    
    \bottomrule
  \end{tabular}
  \caption{HP-ViT performance after fine-tuning with RLE loss and applied TTA during inference}
  \label{tab:ablation_test}
\end{table}


\section{Conclusion}
\label{sec:conclusion}

We introduce HP-ViT, built on ViT backbone and Transformer head for hand pose estimation. Our top-performing model, ViT-Huge, was trained using MPJPE loss and fine-tuned with RLE loss for the last 20 epochs. The ensembling model with different setting contributed to decrease the overall error. Our next step is to extend the pose estimation approach from single images to sequences and integrate motion information to enhance performance.

{
    \small
    \bibliographystyle{ieeenat_fullname}
    \bibliography{main}
}


\end{document}